%% file: main.tex
\documentclass[conference]{IEEEtran}
\IEEEoverridecommandlockouts

\usepackage{cite}
\usepackage{amsmath,amssymb,amsfonts}
\usepackage{algorithmic}
\usepackage{graphicx}
\usepackage{textcomp}
\usepackage{xcolor}
\usepackage{multirow}
\usepackage{amsthm}
\usepackage{mathrsfs}
\usepackage{manyfoot}
\usepackage{booktabs}
\usepackage{algorithm}
\usepackage{listings}
\usepackage{times}
\usepackage{epsfig}
\usepackage[export]{adjustbox}
\usepackage{tabularx}

\def\BibTeX{{\rm B\kern-.05em{\sc i\kern-.025em b}\kern-.08em
    T\kern-.1667em\lower.7ex\hbox{E}\kern-.125emX}}

\begin{document}

\title{ZoDIAC: Zoneup Dropout Injection Attention Calculation\\
}

\author{\IEEEauthorblockN{1\textsuperscript{st} Zanyar Zohourianshahzadi}
\IEEEauthorblockA{\textit{Department of Computer Science} \\
\textit{University of Colorado Colorado Springs}\\
Colorado Springs, Colorado, USA \\
zzohouri@uccs.edu}
\and
\IEEEauthorblockN{2\textsuperscript{nd} Terrance E. Boult}
\IEEEauthorblockA{\textit{Department of Computer Science} \\
\textit{University of Colorado Colorado Springs}\\
Colorado Springs, Colorado, USA \\
tboult@uccs.edu}
\and
\IEEEauthorblockN{3\textsuperscript{rd} Jugal K. Kalita}
\IEEEauthorblockA{\textit{Department of Computer Science} \\
\textit{University of Colorado Colorado Springs}\\
Colorado Springs, Colorado, USA \\
jkalita@uccs.edu}
}

\maketitle

\begin{abstract}
  In the past few years the transformer model has been utilized for a variety of tasks
  such as image captioning, image classification 
  natural language generation, and natural language understanding.
  As a key component of the transformer model, self-attention calculates the attention values by mapping the relationships among the head elements of the source and target sequence,
  yet there is no explicit mechanism to refine and intensify the attention values with respect to the context of the input and target sequences.
  Based on this intuition, we introduce a novel refine and intensify attention mechanism that is called Zoneup Dropout Injection Attention Calculation (ZoDIAC), 
  in which the intensities of attention values in the elements of the input source and target sequences
  are first refined using GELU and dropout and then intensified using a proposed zoneup process which includes the injection of a learned scalar factor. 
  Our extensive experiments show that ZoDIAC achieves statistically significant higher scores under all image captioning metrics using
  various feature extractors in comparison to the conventional self-attention module in the transformer model on the MS-COCO dataset.
  Our proposed ZoDIAC attention modules can be used as a drop-in replacement for the attention components in all transformer models.
  The code for our experiments is publicly available at: https://github.com/zanyarz/zodiac
\end{abstract}


\input{sec1.tex}

\input{sec2.tex}

\input{sec3.tex}
\input{sec4.tex}
\input{sec5.tex}
\input{sec6.tex}
\input{sec7}

\bibliographystyle{IEEEtran}
\bibliography{ALL}

\vspace{12pt}
\color{red}

\end{document}

%% file: sec1.tex
\section{Introduction}\label{sec1}
This work is the extended version of our paper published in IEEE AIxSET 2024. In this version we perform further ablation studies via utlization of convolutional features instead of attention-based features.

For more information and previously published material please refer to ZoDIAC \cite{ZZ24}.

A comprehensive review of methods that deploy attention mechanisms both with LSTMs and in the context of transformers has already revealed that transformer-based \cite{ZZ24,ZZ25,Cornia_2020_CVPR,clipcap,CLIP} attention outperforms the other forms of attention that are utilized with LSTMs and CNNs \cite{ZZ22}. Among all methods that use attention with LSTM as visual attention, they all use soft attention mechanisms \cite{ZZ20,ZZ21,Lu2018NBT,anderson-etal-2017-guided,Anderson2017up-down} rather than hard attention \cite{Xu}.

\par 
From the early days of research on neural networks developed for perception \cite{Rosenblatt}
dating back to at least half a century ago until today, the era of modern neural networks that solely
rely on the self-attention mechanism and feed-forward linear layers in the transformer model \cite{NIPS2017_7181},
the subject of visual attention \cite{Koch1987,ITTI2000} has been a pinpoint of interest for researchers.
\begin{figure}[ht!]
    \centering
    \includegraphics[width=\linewidth]{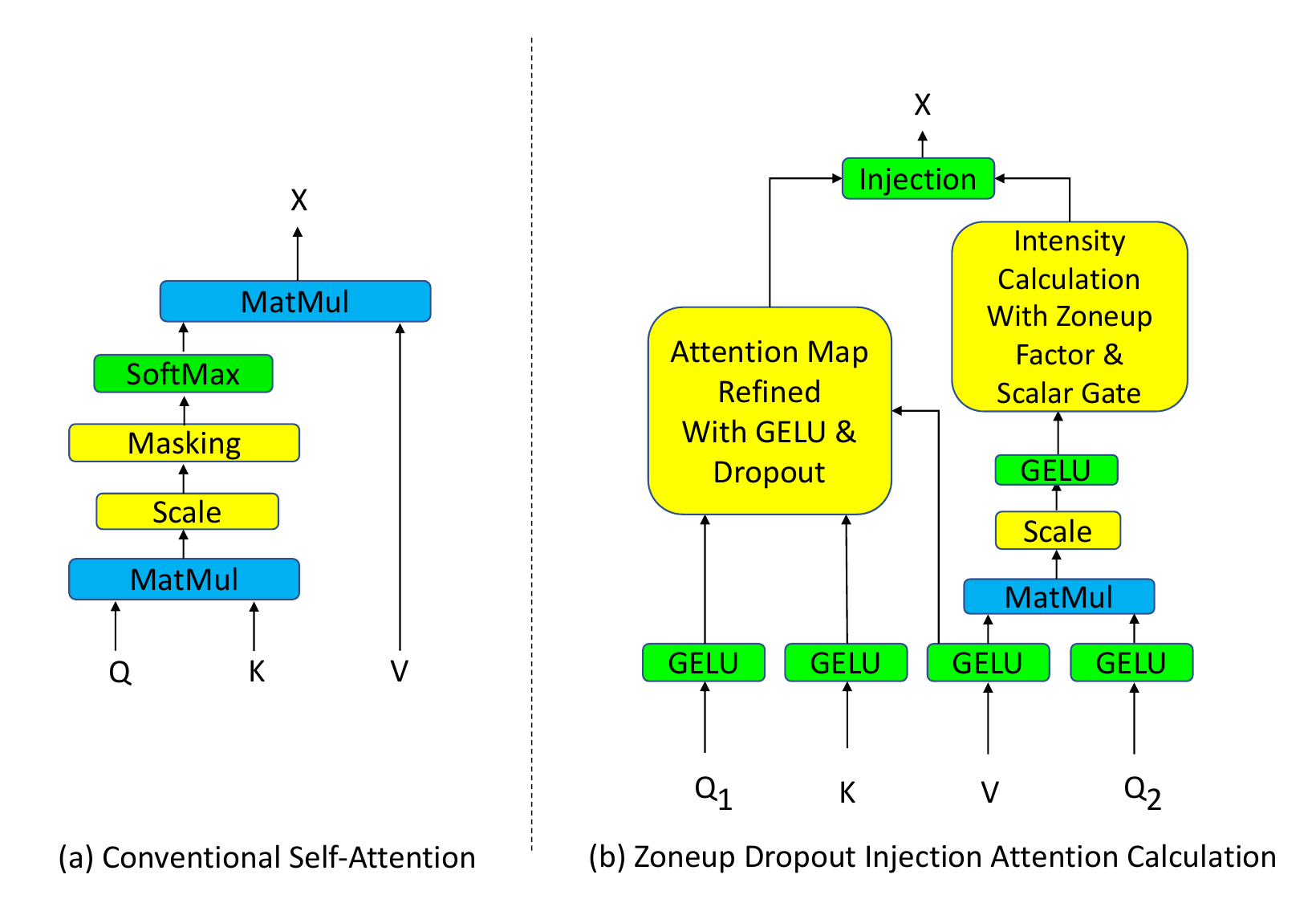}
    \caption{(a) Conventional self-attention used in Transformer \cite{NIPS2017_7181}.
    (b) Our proposed Zoneup Dropout Injection Attention Calculation module.
    ZoDIAC involves two main steps, first refinement of attention map values and then injection of learned scalar intensity value
    using a non-linear projection of queries as secondary queries and sigmoid or tanh as activation functions added with a constant zoneup scalar value.
    The refinement in ZoDIAC involves leveraging GELU \cite{GELU} and dropout \cite{Hinton2014_dropout} non-linear activations.
    For scalar intensity value calculation, we create a secondary attention map from the secondary queries and values.
    Note that GELU is used before and after the matrix multiplication in ZoDIAC on the right side of (b).
    After the intensity calculation, the zoneup added with intensity value is injected (multiplied) element-wise to all attention map values with refined intensities.}
    \label{fig1}
\end{figure}
However, for many years it remained a mystery how to leverage the attention mechanism effectively
end-to-end (without convolutions or recurrences) for various modalities of information.
With the advent of the transformer model \cite{NIPS2017_7181}, the true power of the attention mechanism was revealed to everyone
as this model only relies on linear layers and attention.
By leveraging the encoder of the transformer model and stacking more encoder layers, BERT \cite{BERT}
and other similar models such as DistilBERT \cite{DistilBERT}, RoBERTa\cite{Roberta} and EFL \cite{EFL},
achieved state-of-the-art results in natural language understanding tasks on benchmarks such as SQUAD \cite{SQUAD}, GLEU \cite{GLEU}, Super-GLEU \cite{SUPERGLEU} and SNLI \cite{SNLI}.
In natural language generation, by stacking more transformer decoders GPT-3 \cite{GPT3} has achieved state-of-the-art results using benchmarks such as GLEU and Super-GLEU.
In computer vision, models such as ViT \cite{ViT}, Swin \cite{Liu_2021_ICCV} and DeiT \cite{DEIT} have achieved state-of-the-art results
in image classification on ImageNet-1K \cite{imagenet_cvpr09}.
More recently, CoCa \cite{yu2022coca} and ModelSoups \cite{wortsman22a} pre-trained with text-to-image and image-to-text (vision-language) tasks
have achieved state-of-the-art results.
Directly related to our work and as a generative vision-language task, in image captioning on MS-COCO dataset \cite{MS_COCO} using
Karpathy's test split \cite{Karpathy_2015_CVPR}, models such as mPLUG \cite{MPLUG}, OFA \cite{wang2022OFA}, SimVLM \cite{wang2022simvlm}
have achieved the highest CIDEr \cite{CIDER} scores as well as achieving state-of-the-art results in visual question answering on VQA dataset \cite{VQA} and visual entailment on SNLI-VE \cite{SNLI-VE}.
The important point to note here is that these methods all commonly use the attention components as identically as they were
introduced in the transformer model. Another recent popular practice is to pretrain the transformer model
with huge amounts of parameters and image-text pairs and various forms of contrastive loss as in CoCa and CLIP \cite{CLIP} or rather leverage multi-tasking as in OFA.
It has become evident that the multi-head attention module and self-attention mechanism are useful and interesting tools that deserve more attention and investigation for improvement.
\par
Self-attention maps the internal interactions and relationships among the elements of queries, keys and values of the source and target sequence.
This mapping includes
the application of Softmax on attention map values (which are the resulting values of matrix multiplication over queries and keys) over the values of the source and target sequence.
However, this mapping that includes linear projection of input and target sequences into queries, keys and values followed by matrix multiplication and Softmax leads to a poor attention mechanism if the attention map values are
not well separated (might have values close to each other).
Consider that when communicating with each other and trying to remember and describe a visual event (or object) we saw earlier to someone else,
we tend to describe the most important features (with the highest intensities).
For instance, suppose we have a "brown bear" and a "black bear" in an image and the caption is "a brown bear and a black bear are hunting for fishes in the river".
Here in the given caption the word "bear" might be represented with similar values for both "black bear" and "brown bear" in the attention map. Here "bear" is mapped to both "black bear" and "brown bear" that refer to two different regions of the image.
A linear projection followed by matrix multiplication and softmax for "bear" would fail to capture the difference between "black bear" and "brown bear" and would map "bear" to both "black bear" and "brown bear" in the attention map.
Via attention refinement and intensity value injection, we propose to increase the separation of attention map values for each word in the sentence with respect to the contextual relationships with other words.
In our proposed refinement and intensity injection strategy,
first we refine the attention map values using GELU \cite{GELU} (which always outputs smaller value than the input, where small values are penalized much more than large values, i.e. closer to 1) and dropout \cite{Hinton2014_dropout} (which ensures robustness by randomly eliminating some values),
and then we inject a learned scalar intensity value via element wise multiplication.
This leads to a novel attention mechanism, which we refer to as Zoneup Dropout Injection Attention Calculation (ZoDIAC),
which is displayed in Figure \ref{fig1}(b) and that leverages intensity calculation over the words in the sentence based on the context of each word in the sentence.
Our contributions can be summarized as the following:
\begin{itemize}
    \item We introduce a novel attention mechanism called ZoDIAC that refines and intensifies the attention values for improved mapping between source and target sequences in the self-attention mechanism and our novel ZoDIAC multi-head attention module creates a secondary linear projection of queries for use in intensity value calculation.
    \item Via statistical t-test analysis, we show that ZoDIAC outperforms the conventional self-attention employed in the Transformer model with statistical significance under CIDEr and BLEU scores regardless of the kind of feature extractor used.
    \item We propose the use of an extra linear layer for calculating a single scalar value that is used for intensity injection in the attention values after refinement. This novel strategy leads to higher quality captions generated for input images in image captioning task.
    \item For intensity value calculation and injection we propose a novel regional attention pooling module that creates a leanrned scalar value by averaging the attention values.
    Additionally, for the first time we show that different values for dropout rate used in attention mechanism and other parts of the model successfully creates a refinement effect over attention values.
\end{itemize}
\par

%% file: sec2.tex
\section{Related Work}\label{sec2}
\par
Predating the deep neural networks, template-based methods \cite{Kulkarni,Farhadi_2010,Li_2011} used prior knowledge of visual features and relied on visual feature engineering for visual feature extraction.
The first deep learning models used for image captioning end-to-end, such as the Show and Tell
introduced by Vinyals et al. \cite{Vinyals_2015_CVPR}, and the first attentive deep model, Show, Attend and Tell, introduced by Xu et al. \cite{Xu},
performed better than template-based methods by employing CNNs for visual feature extraction.
Early deep learning methods for image captioning used convolutional architectures
that operate upon the entire image to extract the visual
features in the encoder part of the model \cite{Karpathy_2014,Donahue_2015_CVPR,Xu,Xu2}.
Via leveraging an object detector, Anderson et al. \cite{Anderson2017up-down} introduced bottom-up and top-down attention for image captioning and visual question answering.
In bottom-up attention, the input image is passed through an object detector, usually
Faster-RCNN \cite{FasterRCNN} pre-trained on Visual-Genome dataset \cite{VGenome} using a ResNet-101 \cite{Resnet} convolutional network
pre-trained on ImageNet-1k \cite{imagenet_cvpr09}. From the RoI Align layer in the object detector,
we get the region proposals or the coordinates of the objects in the input image.
This way the model only attends to salient regions in the image.
Visual features are extracted from the detected regions using a CNN backbone,
and these features concatenated with the word embedding at each time step are 
sent to a visual attention LSTM and then to an attention network that performs visual attention.
The attention values and the hypothesis vector of the visual attention LSTM are then sent
to a language LSTM for generating token embeddings at word level \cite{Anderson2017up-down}.
\par
The advent of bottom-up attention enabled the utilization of the transformer model and variations of it for image captioning.
Yu et al. \cite{Yu_2019_mmt}
introduced a Multi-modal Transformer that used multiple views of
the object proposal sets with different orders to provide the encoder in the transformer with different sets of Bottom-up features.
Liu et al. \cite{Liu_2019} similarly employed visual attention values alongside the context attention values and attributes attention values
as cross-modal information.
Li et al. \cite{Li_2019_ICCV} introduced the 
Entangled Transformer by applying weighting on a meshed network of linear transformations of
the queries and values in the visual encoder and the attributes encoder.
Pan et al. \cite{Pan_2020_CVPR} introduced a bilinear pooling mechanism in the conventional self-attention block, resulting in
the x-linear attention block \cite{Pan_2020_CVPR}, which exploits the spatial and
channel-wise bilinear attention values to reveal the second and infinity order
interactions between the multi(or single)-modal input features \cite{Pan_2020_CVPR}.
Guo et al. \cite{Guo_2020_CVPR} created the normalized and geometry-aware self-attention block that exploits geometrical
information presented in visual features.
Cornia et al. \cite{Cornia_2020_CVPR} introduced Meshed-Memory Transformer that  performs linear transformation and sigmoidal gating over the memory states of
encoders in the encoder stack and decoders in the decoder stack of transformer.
Herade et al. \cite{Herade_2019_NIPS} used object labels as attributes
to be concatenated with visual features as input information for the encoder in Transformer.
\par
Recently, we have witnessed the effectiveness of pre-training transformers for image captioning and other tasks.
Specifically, Unified Vision-Language Pre-training \cite{Zhou_VLP} opened the door to pre-training a transformer on vision-language tasks.
More recently, OFA \cite{wang2022OFA} leveraged large-scale vision-language multi-task pre-training to obtain state-of-the-art results on image captioning and other vision and language tasks.
In particular, they pre-train a transformer on object detection (bottom-up feature extraction) and image reconstruction as vision tasks and text infilling as the language pre-training task. 
Another popular trend is to leverage large-scale vision-language pre-training data with transformer models with more parameters such as LEMON \cite{LEMON}, SimVLM \cite{wang2022simvlm}
and mPLUG \cite{MPLUG}, which also leverages the skip connections method inside the encoder and decoder parts of the transformer model.
Another recently proposed method is to pre-train the models with large-scale vision-language data and various forms of contrastive loss as in
CoCa \cite{yu2022coca} and CLIP \cite{CLIP}.
\par
Except for the x-linear attention model \cite{Pan_2020_CVPR} that stays the closets work to ours, all other methods we mentioned in this section
have used the self-attention mechanism identically as it was introduced in the transformer model, 
also none of the modifications in related work we mentioned here address the issue of refining and intensifying the attention values in the source and target sequence.
\par

%% file: sec3.tex
\section{Methodology}\label{sec3}
\par
To understand how ZoDIAC (Firgure \ref{fig1}(b)) leverages the refine and intensify strategy
based on the intuitions we discussed earlier, 
we need to have a detailed understanding of how self-attention (Figure \ref{fig1}(a)) works.
\par
Self-attention is defined as a function of mapping queries and a set of key-value pairs to an output.
A multi-head attention module that includes multiple self-attention heads
is used inside the encoder and decoder parts of the transformer model.
At each stack level in the transformer model, the multi-head attention is used inside the encoder once and the decoder twice.
In the encoder and decoder, the self-attention is used over the input and output
sequences respectively to capture the internal relationships
among the head element of the sequences.
The encoder multi-head attention module operates on the source mask and the decoder multi-head attention module operates on the target mask,
and the cross multi-head attention in the decoder operates on the source mask and the output of encoder multi-head attention and decoder
target masked multi-head attention.
\par
There is a 2-step process for calculating final attention values in self-attention.
The first step entails creating an attention map via the application of the softmax function over the result of matrix multiplication (MatMul) operation
between queries and keys and scaling (and masking while training) over the results before the application of the softmax function.
In the second step, the query-key attention map and the value are used inside a MatMul operation for final attention values generation.
The attention values are later used inside a feed-forward log-softmax layer for word generation at the token level.
\par
We redefine attention as mapping queries and a set of key-value pairs and mapping secondary query and value pairs to an output.
First, we calculate refined attention map values by applying GELU \cite{GELU} and dropout \cite{Hinton2014_dropout} activations.
Specifically, the dropout layer used for attention refinement in ZoDIAC is set with a probability of 0.2 rather than 0.1 for other parts of the model.
Combining this increase in dropout rate in ZoDIAC attention mechanism with the application of GELU we achieve our refinement effect for
our refine and intensify strategy.
The secondary query and value pairs are used for calculating an intensity map that is not softmax activated and is
averaged (multiplied by mask values while training) and sent to an activation function that can produce a scalar value, in this case we choose sigmoid and tanh.
After injection of scalar intensity value into the attention map values, we achieve attention values with refined intensities, which
are going to be used in a feed-forward log-softmax layer for word generation at the token level.
Via the application of GELU and dropout with an increased rate compared to other parts of the model (0.2 compared to 0.1)
and generating an intensity map and an intensity value from it and applying the intensity value to the refined attention map values,
ZoDIAC (Figure \ref{fig1}(b)) resembles how information should be refined and intensified to achieve improved memorization (learning) and generalization (inference).
\par
\begin{figure*}[ht!]\centering
    \includegraphics[width=8.6cm]{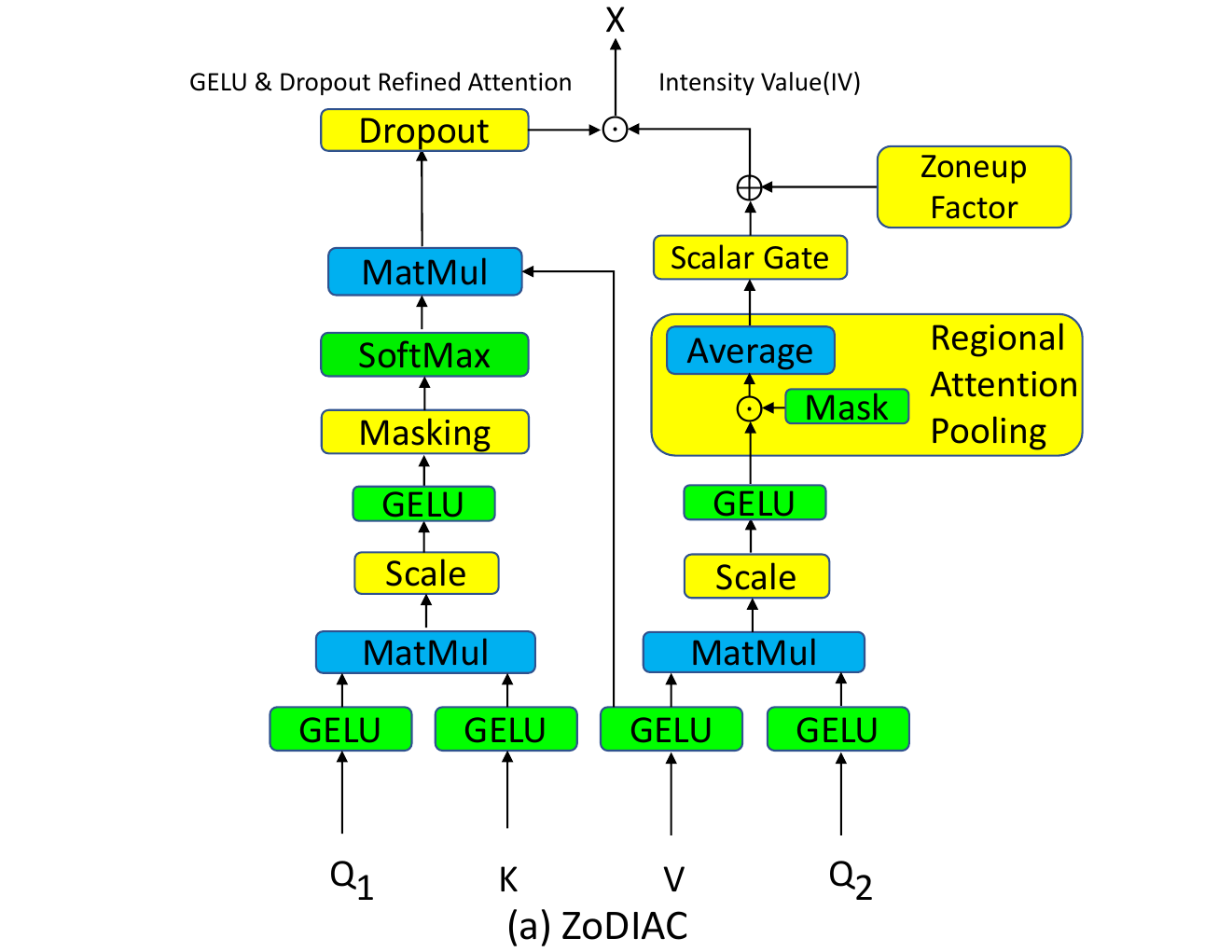}
    \includegraphics[width=8.6cm]{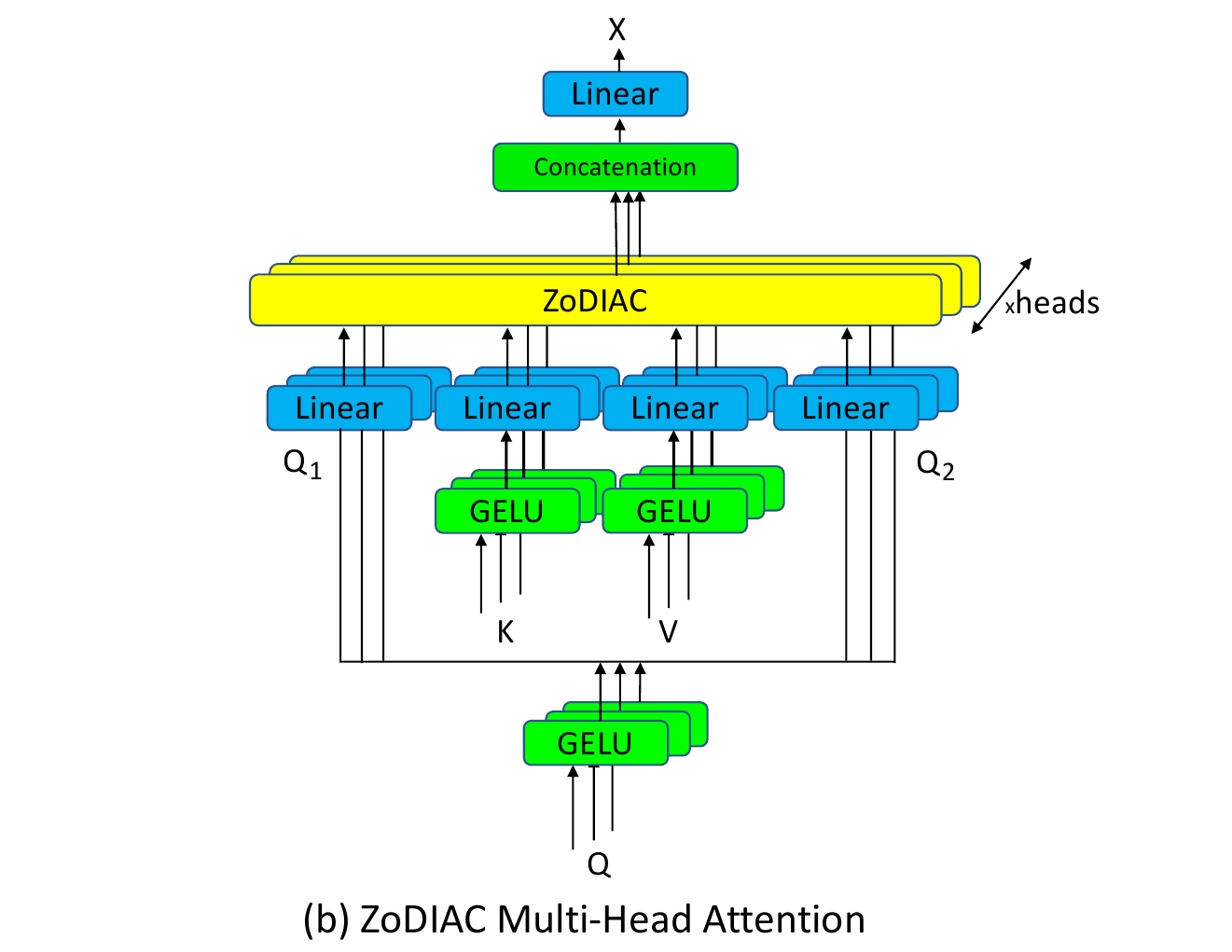}
    \caption{(a) ZoDIAC: RA in left, and IV in right side. (b) ZoDIAC Multi-Head Attention: Augmented with GELU and linear layer.}
    \label{fig2}
\end{figure*}
\par
\subsection{GELU and Dropout for Refinement}
\par
Gaussian Error Linear Units (GELU) \cite{GELU} non-linear function combines the properties of
zoneout \cite{Zoneout}, dropout \cite{Hinton2014_dropout} and ReLu activation, all in one place.
ReLu and dropout are like each other because they both block some incoming signals from the previous layer.
Their difference is that ReLu drops the signals that have values below zero and on the other hand,
dropout drops (sets to zero) some portions of the signal randomly based on the given dropout rate.
zoneout \cite{Zoneout} is the complete opposite of dropout because some portion of the signals are kept identical (multiplied by one) and
are selected to be identical to the previous layer.
\par
Combining all these features, GELU activation function is defined (in a simplified form) as the following:
\par
\begin{equation}\label{eq1}
    GELU(x) = \frac{x}{2}+\frac{x}{\sqrt{\pi}}\int_{0}^{\frac{x}{\sqrt{2}}} e^{-t^2} dt .
\end{equation}
\par
Looking at Eq.(\ref{eq1}), we realize that the GELU function first cuts the signal in half,
then it searches for the right amount of signal to be added to the previously cut signal.
Euler's number ($e$) is used in the form of $e^{-t^2}$ to create a curve for Area Under Curve (AUC) calculation, which correlates
with the correct amount of signal that should be added to the previously cut signal at each time step $t$. Therefore, Eq.(\ref{eq1})
shows how the intensity of input signals (element-wise) is being refined by the GELU activation in Figure \ref{fig2}.
GELU can also be approximated with $x\sigma(1.702x)$. The sigmoid function ($\sigma$) generates a value between 0 and 1.
Therefore, multiplying the result of the sigmoid function by the incoming signal ($x$)
always guarantees a refinement (lowering) effect over the input signal.
\par
By utilizing an increased dropout rate of 0.2 in comparison with 0.1 used for all other parts of our ZoDIAC-Transformer,
we achieve our goal of creating another refinement effect by allowing the model to randomly learn which attention map values
should be set to zero (eliminated).
Combining the application of GELU and dropout non-linear functions
we achieve the refinement effect we seek for our refine and intensify strategy on the attention map values.
\par
\subsection{Refined Attention (RA)}
\par
In order to create the refined attention map values and intensity value before injection in Figure \ref{fig2}(a), we leverage
parallel attention and intensity map calculation.
Specifically, as shown in Figure \ref{fig2} (b), input sequence is transformed into secondary ($Q_2$), primary ($Q_1$) queries, keys ($K$) and values ($V$). 
Via the application of MatMul operation over the primary query ($Q_1$) and keys ($K$) we calculate the attention map.
The utilization of GELU over the primary query and key and the result of MatMul creates an intensity refinement effect
over the attention map before the application of Softmax over the attention map.
The attention map and the values ($V$) are used inside MatMul operation
for attention map value generation. The attention map values are then sent to a dropout layer with a particular dropout rate that is different than
the dropout rate used in other parts of the model.
We refer to the dropout rate used in other parts of the ZoDIAC-Transformer model as the system dropout rate and the dropout rate used in ZoDIAC attention mechanism for
attention map values refinement as ZoDIAC dropout rate.
Using a different dropout rate inside our model forces it to randomly
learn and eliminate the least important parts of the sequence. This is shown on the left side of Figure \ref{fig2}(a).
With masking operation identical to how it is performed in self-attention \cite{NIPS2017_7181} (Figure \ref{fig1}(a)),
dimension of key denoted as $d_k$, the dropout rate for ZoDIAC attention mechanism in RA denoted as $\delta_Z$ and GELU denoted as $G$ and dropout as $D$,
and MatMul operation not shown for conciseness,
the refined attention (RA) depicted in left side of Figure \ref{fig2}(a) is put together as the following in Eq.(\ref{eq3}-\ref{eq2}):
\par
\begin{equation}\label{eq3}
    RA_{map}(Q,K) = Softmax(G\big(\frac{G(Q) G(K^T)}{\sqrt{d_k}}\big)).
\end{equation}
\begin{equation}\label{eq2}
    RA(Q,K,V) = D_{\delta_Z}\big(RA_{map}(Q,K) G(V)\big).
\end{equation}
\par
\subsection{Intensity Value (IV)}
\par
Via performing MatMul operation over secondary query ($Q_2$) and value, we create an intensity
map. The computation process of this intensity map is identical to how the refined attention map is calculated in Eq.(\ref{eq3}), except that
instead of using the primary query and key for intensity map generation, here we use secondary query and value without using softmax activation for this purpose.
After the intensity map is computed, the boolean mask (with the same dimensions intensity map) is multiplied
element-wise to the intensity map. After the injection of the mask into the intensity map, via performing an average
calculation operation over all the remaining attention map values in the intensity map, we achieve a Regional Attention Pooling (RAP) scalar value.
Via applying a scalar (element-wise) activation function of choice
over the RAP scalar value, we achieve the intensity value.
By adding a proposed zoneup factor into the intensity value we achieve the final intensity value,
this value is then injected (element-wise) into the refined attention map values as we explain in the following section.
\par
Considering the intensity value calculation function as IV, regional attention pooling function as RAP,
mask as the same boolean mask used for masking operation in RA,
the zoneup factor as $\zeta$, the scalar activation gate as $\phi$, dimension of
value as $d_v$, total number of values in intensity map (or mask) as $N$, element-wise addition and multiplication as $\oplus$ and $\odot$,
and GELU denoted as $G$, the intensity value calculation 
in the right side of Figure \ref{fig2}(a)) is shown in the following equations Eq.(\ref{eq4}-\ref{eq8}):
\par
\begin{equation}\label{eq4}
    \textstyle{IV_{map}(Q,V) = G\big(\frac{G(Q) G(V^T)}{\sqrt{d_v}}\big).}
\end{equation}
\begin{equation}\label{eq5}
    \textstyle{IV_{pool}(Q,V,mask) = IV_{map}(Q,V)\odot mask.}
\end{equation}
\begin{equation}\label{eq6}
    \textstyle{RAP(IV_{map},none)=\frac{\sum_{i=1}^{N} IV_{map}(Q,V)_{(i)}}{N}.}
\end{equation}
\begin{equation}\label{eq7}
    \textstyle{RAP(IV_{map},mask) = \frac{\sum_{i=1}^{N} IV_{pool}(Q,V,mask)_{(i)}}{\sum_{i=1}^{N} mask(i)}.}
\end{equation}
\begin{equation}\label{eq8}
    \textstyle{IV(Q,V) = \zeta \oplus \phi \big(RAP(IV_{map},mask)\big).}
\end{equation}
\par
When intensity value is used inside the decoder, Eq.(\ref{eq7}) is infused into Eq.(\ref{eq8}).
When used inside the encoder, where there is no masking, Eq.(\ref{eq6}) is injected into Eq.(\ref{eq8}).
The scalar activation gate could be sigmoid or tanh ($\phi=\sigma $ or $tanh$).
Note the difference between masking in RA
and the injection of the mask here in RAP, which eliminates worthless connections (setting to $0$), and leaves
the remaining connections for use in RAP in IV.
When RA and self-attention \cite{NIPS2017_7181} are used inside the decoder, to prevent leftward
tokens to attend over the future (rightward) tokens in the target sequence, masking is performed.
This masking includes masking out all values (future positions) in the attention map that the model should not attend to while training.
This is the reason behind naming RAP in IV.
\par
\subsection{ZoDIAC Attention Mechanism}
\par
With refined attention calculation function denoted as RA, intensity value calculation function as IV,
input sequences (with positional encoding information added \cite{NIPS2017_7181}) transformed via linear layers into primary query denoted as
$Q_1$, secondary query as $Q_2$, key as $K$ and value as $V$, and element-wise multiplication as $\odot$,
we define ZoDIAC (Figure \ref{fig2}(a)) as the following:
\par
\begin{equation}\label{eq9}
    \textstyle{ZoDIAC(Q_1,Q_2,K,V) = RA(Q_1,K,V) \odot IV(Q_2,V).}
\end{equation}
\par
As briefly mentioned earlier, ZoDIAC is used inside a modified version of multi-head attention that can support ZoDIAC with necessary
sources of information. This is further explained in detail in the following section.
\par
\subsection{ZoDIAC Multi-Head Attention (ZMHA)}
\par
In the Transformer model, the input (source and target) sequence is augmented with positional encoding information \cite{NIPS2017_7181}.
After the positional information is added to the sequence, the sequence is fed to linear layers that transform the input sequence into query ($Q$), key ($K$) and value ($V$).
The query, key and value are used inside self-attention modules as separate heads inside the multi-head attention module.
We redefined self-attention earlier and constructed the ZoDIAC attention mechanism, therefore,
we need to modify (redefine) multi-head attention in a way that enables this module to support
the employment of ZoDIAC attention mechanism as separate attention heads.
Our first modification involves adding a linear layer inside our modified multi-head attention
that transforms the input sequence into a secondary query ($Q_2$).
The second modification involves adding the GELU activation for intensity refinement and increased non-linearity.
This is because we need to refine the intensities of values, keys and secondary and primary queries before the linear layers that transform them.
\par
Considering concatenation defined as brackets, the total number of heads denoted as $h$, each ZoDIAC head az $Z$, GELU denoted as $G$,
the first transformation of query as primary query denoted as $Q_1$, the second transformation of query as secondary query denoted as $Q_2$,
last linear layer that transforms the attention values denoted as $W_o \in \mathbb{R}^{hd_k \times d_{\text{model}}}$ (as shown in Eq.(\ref{eq12})),
$x$ as a variable that could be replaced with the secondary query, primary query, key and value,
and $W_{x} \in \mathbb{R}^{d_{\text{model}} \times d_x}$ (as shown in Eq.(\ref{eq10})) as a stereotype for other linear layers that transform the key, value and secondary and primary queries,
the ZoDIAC-Multi-Head Attention (ZMHA) shown in Figure \ref{fig2}(b) can be put together as in the following equations Eq.(\ref{eq10}-\ref{eq12}):
\begin{equation}\label{eq10}
    {x_{(g)}=W_x\big(G(x)\big).}
\end{equation}
\begin{equation}\label{eq11}
    Z_i = ZoDIAC(Q_{1(g)},Q_{2(g)},K_{(g)},V_{(g)}).
\end{equation}
\begin{equation}\label{eq12}
    {ZMHA(Q_1,Q_2,K,V)=W_o[Z_1,Z_2,...,Z_h].}
\end{equation}
\par

%% file: sec4.tex
\section{Experiments and Results}\label{sec5}
\subsection{Implementation Details} \label{sec3.6}
\textbf{MS-COCO:} This is the most commonly used dataset among researchers for image captioning and object detection.
Due to the huge size of this dataset, standard splits have been defined for training, testing and validation purposes.
We use the standard MS-COCO \cite{MS_COCO} training and validation sets for
training phase and we use the Karpathy's test split \cite{DenseCap_2016} for inference and testing phase,
which is the most commonly used test split for image captioning models.
\par
\textbf{Metrics:} We perform our experiments using common image captioning metrics such as
BLEU \cite{BLEU_2002}, METEOR \cite{METEOR}, ROUGE \cite{ROUGE}, CIDER \cite{CIDER} and SPICE \cite{SPICE}.
Employing metrics that focus on the textual features of the generated captions and ground truth captions,
such as BLEU \cite{BLEU_2002}, ROUGE \cite{ROUGE} or METEOR \cite{METEOR},
ensures the quality of generated captions from a neural machine translation vantage.
On the other hand, employing metrics that consider the quality of vision-language relationships such as CIDER \cite{CIDER} and SPICE \cite{SPICE}
ensures the quality of generated captions from both neural machine translation and computer vision vantages.
The CIDER metric \cite{CIDER},
reflects the quality of relationships between textual and visual properties of input image and input ground truth captions.
This is achieved by computing Term Frequency and Inverse Document Frequency (TF-IDF) and considering the level of relevancy between a set of captions and the input image.
The SPICE metric \cite{SPICE} considers spatial relationships among objects in the image using
scene graphs and then evaluates the quality of captions based on the existence of the discovered spatial relationships in the captions.
\par
\textbf{Image Representation:}
For object detections and region proposals we use Faster-RCNN \cite{FasterRCNN} pre-trained on Visual Genome \cite{VGenome} with ResNet101 pre-trained on ImageNet \cite{imagenet_cvpr09}.
The bounding box coordinates detected by Faster-RCNN are used for feature extraction using a variety of methods in our experiments.
Specifically, we use CLIP \cite{CLIP} models pre-trained with contrastive loss on huge amounts of image-text pairs using datasets such as
ImageNet \cite{imagenet_cvpr09}, YFCC100M \cite{YFCC100M_2016}, OpenImages \cite{OpenImages}, Conceptual Captions \cite{Sharma_etal_2018} and SBU Captions \cite{SBU}.
The models include ViT \cite{ViT} models such as ViT-B32, ViT-L14, ViT-L14@336px and ResNet \cite{Resnet} models such as ResNet-101 and ResNet-50.
We also use models such as BEiT \cite{BEiT} and ConvNeXt \cite{convnext}.
These models are both pre-trained with cross-entropy loss on ImageNet \cite{imagenet_cvpr09} but BEiT is a purely attention-based and ConvNeXt is a convolutional model.
\par
\textbf{Text Representation:}
In our experiments, we use learned embeddings from scratch instead of employing Word2Vec \cite{Word2vec},
Glove \cite{Glove} or BERT \cite{BERT} embeddings. Our model has to learn representations from scratch,
instead of using an external source of knowledge,
resulting in a fair and robust comparison between our proposed ZoDIAC-Transformer and Transformer model.
\par
\textbf{Loss, Optimizer \& Learning Rate:}
We use the same training and evaluation configurations for all models in our experiments.
Specifically, we train the models using cross-entropy loss and the Adam optimizer \cite{Adam}, specifically we use a modified version of this optimizer introduced in the
transformer model called the Noam optimizer \cite{NIPS2017_7181}.
We train all models on $5$ GeForce GTX 1080 Ti cards separately,
with a learning rate of $5e-4$ and learning rate decay of $0.8$ every $3$ epochs, with a batch size of $20$ for a maximum of $30$ epochs.
When tanh is used as a scalar activation gate ($\phi=tanh$), we train the ZoDIAC-Transformer only for $20$ epochs.
\par
\textbf{ZoDIAC-Transformer and Transformer Configs:}
In all of our experiments, we utilize a stack of $6$ encoders and $6$ decoders, and we use $8$ heads in each encoder and decoder for both ZoDIAC-Transformer and Transformer model.
Similarly, other model parameters such as the dimensions of primary query, secondary query, key and value (which should be identical) and feed-forward layers are identical and are set to $512$ and $2048$ respectively.
For finding the optimal zoneup factor in our ablation studies, we use a stack of $5$ encoders and $5$ decoders alongside investigating the effectiveness of other parts of ZoDIAC.
\par
\textbf{Hyper Parameters:} We found a system dropout rate of $0.1$, a ZoDIAC dropout rate ($\delta_Z$) of $0.2$, and
a zoneup factor ($\zeta$) of $1$ as the best configuration for ZoDIAC.
We use sigmoid ($\sigma$) and $\tanh$ as scalar activation gate ($\phi$).
\par
\textbf{Batch \& Beam Sizes:}
In all of our experiments we use a batch size of $20$ and for evaluation, we use beam search with a beam size of $3$.
In total we perform $105$ experiments including $5$ runs for $3$ models using $7$ different feature extraction methods.
These models are namely Transformer, ZoDIAC-Transformer with $\phi=tanh$ and ZoDIAC-Transformer with $\phi=\sigma$.
\par
%

\subsection{Discussion and Results}
\par
The results of our experiments for average of 5 runs for each model using a variety of feature extractors are reported in Table \ref{table1}.
We consider the CIDER score as the most important metric for all our experiments, as this metric captures the vision-language relationships in a more general way than SPICE that considers
spatial relationships or certainly than BLEU, METEOR or ROUGE that solely rely on n-gram calculations (textual features).
In a detailed analysis we report the t-test results in the form of p-values for all the results that are higher under any metric for any ZoDIAC model
compared with the transformer model. For this reason in Table \ref{table1} we report the average of 5 runs for each model and
also the p-values for the results that are equal or lower to the results of transformer are not reported.
\par
\begin{table*}[ht]
    \centering
    \caption{Comparison of our proposed ZoDIAC-Transformer (ZTR) and Transformer (TR) trained with cross-entropy optimization using different feature extractors on Karpathy's test split \cite{Karpathy_2015_CVPR}.
    \dag denotes @336px indicating the resolution of the region used for feature extraction, * denotes the use of tanh as the scalar activation gate, and RN denotes ResNet.}
    \label{table1}
    \begin{tabular}{ccccccccc}
        \hline
        \multicolumn{2}{c}{Model} & \multicolumn{1}{c}{Feature extractor} & \multicolumn{6}{c}{Cross-entropy average of 5 runs / p-value} \\ \cline{4-9} 
        \multicolumn{2}{c}{} & {(Backbone)} & B1 & B4 & M & R & C & S \\ \hline
        
        \multicolumn{2}{c}{ZTR} &  CLIP-ViT-L14\dag\cite{CLIP} & \textbf{78.1/1e-18} & \textbf{37.2/1e-02} & 28.7 / -  & \textbf{57.8/2e-02} & \textbf{121.0/5e-04} & \textbf{21.9/1e-02} \\
        \multicolumn{2}{c}{ZTR*} &  CLIP-ViT-L14\dag\cite{CLIP} & \textbf{78.8/4e-37} & \textbf{37.6/1e-02} & 28.3/ -  & \textbf{57.8/}3e-01 & \textbf{121.3/2e-02} & \textbf{21.9/}8e-1 \\
        \multicolumn{2}{c}{TR} &  CLIP-ViT-L14\dag\cite{CLIP} & 77.5/ - & 36.8/ - & 28.7/ - & 57.6/ - & 120.9/ - & 21.8/ - \\ \hline
        
        \multicolumn{2}{c}{ZTR} &  CLIP-ViT-L14 \cite{CLIP} & \textbf{78.0/1e-10} & \textbf{37.2/4e-03} & \textbf{28.7/3e-02} & \textbf{57.7/2e-03} & \textbf{120.0/1e-02} & \textbf{21.9/}8e-01 \\
        \multicolumn{2}{c}{ZTR*} &  CLIP-ViT-L14 \cite{CLIP} & \textbf{78.8/3e-64} & \textbf{37.6/5e-05} & 28.3/ - & \textbf{57.8/1e-05} & \textbf{120.3/1e-02} & \textbf{21.9/}3e-01 \\
        \multicolumn{2}{c}{TR} &  CLIP-ViT-L14 \cite{CLIP} & 76.5/ - & 36.0/ - & 28.6/ - & 57.1/ - & 119.4/ - & 21.8/ - \\ \hline
        
        \multicolumn{2}{c}{ZTR} &  BEiT$_{large}$ \cite{BEiT} & \textbf{77.2/6e-12} & \textbf{36.0/2e-04} & \textbf{28.3/2e-02} & \textbf{56.8/4e-02} & \textbf{118.2/5e-06} & \textbf{21.7/4e-02} \\
        \multicolumn{2}{c}{ZTR*} &  BEiT$_{large}$ \cite{BEiT} & \textbf{77.9/2e-49}& \textbf{36.6/2e-04} & 27.9/ - & \textbf{57.0/1e-04} & \textbf{117.8/2e-03} & 21.4/ -\\
        \multicolumn{2}{c}{TR} &  BEiT$_{large}$ \cite{BEiT} & 76.3/ - & 35.2/ - & 28.2/ - & 56.6/ - & 115.3/ - & 21.5/ - \\ \hline
        
        \multicolumn{2}{c}{ZTR} &  CLIP-ViT-B32 \cite{CLIP} & \textbf{76.7/3e-10} & \textbf{35.6/1e-04} & 27.8/ - & \textbf{56.5/1e-03} & \textbf{115.5/1e-02} & \textbf{21.3/1e-02} \\
        \multicolumn{2}{c}{ZTR*} &  CLIP-ViT-B32 \cite{CLIP} & \textbf{77.3/2e-53}& \textbf{35.9/3e-06} & 27.6/ - & \textbf{56.7/8e-08} & \textbf{115.7/1e-03} & 21.1/ - \\
        \multicolumn{2}{c}{TR} &  CLIP-ViT-B32 \cite{CLIP} & 75.6/ - & 34.7/ - & 27.9/ - & 56.1/ - & 114.3/ - & 21.1/ - \\ \hline
        
        \multicolumn{2}{c}{ZTR} &  CLIP-RN101 \cite{CLIP} & 76.2/ - & \textbf{35.2/2e-05} & 27.7/ - & \textbf{56.3/4e-05}  & \textbf{113.9/3e-02}  & 21.0/ -  \\
        \multicolumn{2}{c}{ZTR*} &  CLIP-RN101 \cite{CLIP} & \textbf{77.1/7e-53} & \textbf{35.8/2e-05} & \textbf{27.3/1e-10} & \textbf{56.5/2e-10}  & \textbf{114.2/8e-03}  & 20.4/ - \\
        \multicolumn{2}{c}{TR} &  CLIP-RN101 \cite{CLIP} & 76.5/ - & 34.5/ - & 27.8/ - & 55.9/ - & 113.1/ - & 21.0/ - \\ \hline
        
        \multicolumn{2}{c}{ZTR} & ConvNeXt$_{large} $\cite{convnext} & \textbf{76.8/1e-41} & \textbf{35.7/3e-14} & \textbf{28.0/1e-19} & \textbf{56.5/2e-05} & \textbf{116.4/1e-18} & \textbf{21.2/1e-04} \\
        \multicolumn{2}{c}{ZTR*} & ConvNeXt$_{large} $\cite{convnext} & \textbf{77.4/8e-87} & \textbf{35.7/1e-09} & 27.5/ - & \textbf{56.4/7e-15} & \textbf{116.1/1e-15} & \textbf{21.1/1e-02} \\
        \multicolumn{2}{c}{TR} & ConvNeXt$_{large} $\cite{convnext} & 75.3/ - & 34.0/ - & 27.7/ - & 55.6/ - & 113.5/ - & 21.0/ - \\ \hline

        \multicolumn{2}{c}{ZTR} &  CLIP-RN50 \cite{CLIP} & \textbf{75.5/9e-04}  & \textbf{34.6/2e-02} & \textbf{27.5/7e-03}  & \textbf{55.8/3e-02}  & \textbf{111.2/2e-02}  & 20.5/ - \\
        \multicolumn{2}{c}{ZTR*} &  CLIP-RN50 \cite{CLIP} & \textbf{76.5/1e-43}  & \textbf{35.0/7e-04} & 27.0/ - & \textbf{56.0/1e-05}  & \textbf{111.4/1e-02}  & 20.4/ - \\
        \multicolumn{2}{c}{TR} &  CLIP-RN50 \cite{CLIP} & 75.1/ - & 34.0/ - & 27.4/ - & 55.6/ - & 110.8/ - & 20.7/ - \\ \hline
    \end{tabular}
\end{table*}
\par
By looking at the results from Table \ref{table1}, we observe that results of ZoDIAC models, both with tanh and sigmoid activations as the scalar gate for intensity value calculation,
outperform the Transformer model under all metrics with statistical significance except for SPICE and METEOR, in which the ZoDIAC models achieve almost the same results.
For statistical significance, we choose a p-value of $5e-2$ and any p-values less than this value are statistically significant and are typed in bold in Table \ref{table1}.
As we observe using all feature extractors except for BEiT and ConvNeXt, when tanh is used as a scalar activation gate instead of the sigmoid ($\sigma$), the results are further improved
under the CIDEr metric.
We believe this is since the tanh activation gate creates a value between negative one and one, whereas the sigmoid function creates a value between 0 and 1,
By adding the resulting values to zoneup factor (set as 1), ZoDIAC is creating different refinement effects.
When tanh is used, the intensity value can be less than 1 or greater than 1.
When sigmoid is used, considering the zoneup factor, the intensity value is always between 1 and 2.
In other words, when tanh is used, the model can decrease or increase the intensity of attention values,
whereas when sigmoid is used the model only has to learn how much it should increase the intensity of attention values.
\par
\par
By looking at the results from Table \ref{table1}, we can conclude that ZoDIAC is a superior attention mechanism for image captioning in
comparison with the conventional self-attention mechanism inside the Transformer model.
This conclusion is based on the observation that ZoDIAC achieves higher CIDEr scores with statistical significance using all feature extractors.
Same applies for BLUE1 and BLEU4 scores. The fact that we don't see the same statistical significance for SPICE, METEOR and ROUGE-L scores
is that these metrics are not as sensitive as CIDEr and BLEU metrics to small changes in the results.
Also this is in indicator that perhaps the zoneup factor used for our experiments is not the optimal value for all feature extractors.
Before we start the experiments using all feature extractors we performed a grid search for the zoneup factor and we found that the best results are achieved
when the zoneup factor is set to 1. These results are reported under the form of ablation studies in the following.
\par
\par
At the current time, the state-of-the-art results in image captioning are generated by models such as OFA \cite{wang2022OFA}, LEMON \cite{LEMON} and SimVLM \cite{wang2022simvlm}
that utilize pre-training with multi-tasking and huge amounts of vision-language data and huge parameters for the transformer model.
We do not perform a comparison with state-of-the-art because we do not leverage pre-training with large vision-language data or multi-tasking
or contrastive learning methods.
Instead, we only leverage cross-entropy loss training for image captioning.
As a potential limitation we only test the ZoDIAC-Transformer model for image captioning with cross-entropy loss training without web-scale pre-training data
and we leave the extension of our work to other tasks such as image classification and natural language processing tasks for future work.
As the main goal in this work, we successfully introduced the ZoDIAC attention mechanism and we showed that it can be used as a drop-in replacement for the self-attention mechanism inside the Transformer model
by creating the ZoDIAC-Transformer model, therefore we did not perform a comparison with state-of-the-art models that leverage pre-training with web-scale data and multi-tasking.
\par

%% file: sec5.tex
\section{Ablations}
To reveal the effectiveness of different parts of ZoDIAC we perform further ablation studies.
\begin{table}[!h]\centering
    \footnotesize
    \caption {Results for ablations grid search. Scalar gate is denoted as S-Gate. System and ZoDIAC dropout rates are denoted
    as S-Dr and Z-Dr. $\textbf{\dag}$ refers to ZTR version without GELU, which achieves worst performance and ZF denotes the Zoneup Factor.}
    \begin{tabular}{ccccccccc}
        \toprule 
        \textbf
        & & & & & & {Metrics} \\
        S-Dr & Z-Dr & ZF & S-Gate & B1 & B4 & S & C \\
        0.05 & 0.1 & 0 & - & 75.41 & 34.26 & 19.77 & 109.88 \\
        0.1 & 0.05 & 0 & - & 75.52 & 34.66& 19.72 & 108.46 \\
        0.1 & 0.2 & 0 & - & 74.01 & 32.79 & 19.62 & 108.45 \\
        0.2 & 0.1 & 0 & - & 75.78 & 34.85 & 19.81 & 109.1 \\
        0.1 & 0.1 & 0 & - & 75.78 & 34.3 & 19.91 & 108.13 \\ 
        0.2 & 0.2 & 0 & - & 75.31 & 34.67 & 20.12 & 109.09 \\
        0.2 & 0.25 & 0 & - & 75.3 & 34.08 & 20.26 & 108.76 \\
        0.2 & 0.15 & 0 & - & 74.98 & 34.03 & 19.78 & 108.35 \\
        \textbf{0.2} & \textbf{0.3} & 0 & - & \textbf{75.59} & \textbf{34.91} & \textbf{20.29} & \textbf{110.57} \\
        \hline
        0.1	& 0.1 & 1.0 & - & 75.49 & 34.86 & 20.32 & 110.7 \\
        0.1	& 0.2 & 1.0 & - & 74.75 & 34.29 & 20.09 & 108.05 \\
        0.1	& 0.3 & 1.0 & - & 75.01 & 33.89 & 19.85 & 107.52 \\
        0.1	& 0.4 & 1.0 & - & 74.89 & 33.84 & 19.97 & 107.69 \\ 
        0.2$\textbf{\dag}$ & 0.3$\textbf{\dag}$ & 1.0 & - & 73.69$\textbf{\dag}$ & 33.12$\textbf{\dag}$ & 19.45$\textbf{\dag}$ & 106.04$\textbf{\dag}$ \\
        \textbf{0.2} & \textbf{0.3} & 1.0 & - & \textbf{75.82} & \textbf{35.34} & \textbf{20.46} & \textbf{111.88} \\
        \hline
        0.1 & 0.2 & 2.0 & $\sigma$ & 75.85 & 27.62 & 20.75 & 112.03 \\
        0.1 & 0.2 & 1.1 & $\sigma$ & 75.79 & 26.52 & 19.94 & 108.56 \\
        0.2 & 0.3 & 1.0 & $\sigma$ & 75.83 & 27.94 & 20.82 & 113.51 \\
        0.1 & 0.1 & 1.0 & $\sigma$ & 76.24 & 27.64  & 20.89 & 113.05 \\ 
        \textbf{0.1} & \textbf{0.2} & 1.0 & $\sigma$ & \textbf{77.01} & \textbf{36.12} & \textbf{21.35} & \textbf{115.53} \\
        \bottomrule 
    \end{tabular}
    \normalsize
    \label{table4}
\end{table}
\par
By looking at Table \ref{table4}, we observe that when there is no scalar activation gate used inside ZoDIAC, a system dropout rate of $0.2$ and a ZoDIAC dropout rate of $0.3$
is the best combination. However, when the scalar activation gate is used, the system dropout rate of $0.1$ and the ZoDIAC dropout rate of $0.2$ is the best combination, as used in our experiments. As a conclusion for Table \ref{table4}, the system dropout rate should be lower than the ZoDIAC dropout rate.
\par
Further, we want to confirm that the performance gain by ZoDIAC are not a consequence of employing more parameters solely.
We do this by extending transformer with more parameters and compare it with ZoDIAC. ZoDIAC has around 59M parameters, whereas the transformer model used with the same
architecture configurations has around 54.5M parameters.
\par
We increase the parameters of transformer by increasing the dimension of output layer (d-ff) of transformer from 2048 to 4096.
This would increase there parameters from 2048x4028 (around 4.2M) to 4096x4096 (around 16.8M), and increase of 12.6M parameters. We refer to this model as transformer+.
\par
We also increase the parameters of transformer by increasing the dimensions of attention layers (dim-model=768, instead of 512).
For each transformer layer we have one encoder multi-head attention and two decoder multi-head attentions.
Instead of 4 linear layers for each multi-head attention module with 512x512 parameters, we have 4 linear layers with 768x768 parameters.
This would increase the parameters from 4x512x512 (around 1M) to 4x768x768 (around 2.4M), an increase of 1.4M parameters for each multi-head attention module.
Considering that there are 6 encoders and 6 decoders, we will have 6+(6x2)=18 multi-head attention modules, and the total increase in parameters would be 18x1.4M=25.2M.
We refer to this model as transformer++.
Table \ref{table:supp2} shows the experiment results for transformer++ and transformer+ compared with transformer and ZoDIAC using CLIP-L14/336px for feature extraction.
We observe that both transformer++ and transformer+ have lower performance compared with transformer and ZoDIAC. As a result, we can conclude that the performance gain in ZoDIAC is not solely due to the use of more number of trainable parameters in the model.
\begin{table}[!ht]\centering
    \footnotesize
    \caption {Results for ablations on number of parameters.}
    \begin{tabular}{ccccccccc}
        \toprule 
        \textbf
        & & & & & & {Metrics} \\
        Model & Params & B1 & B4& M & R & C & S \\
        ZTR & 59.0M & \textbf{78.1} & \textbf{37.2} & 28.7 & \textbf{57.8} & \textbf{121.0} & \textbf{21.9} \\
        TR & 54.5M & 77.5 & 36.8 & 28.7 & 57.6 & 120.9 & 21.8 \\
        TR+ & 67.1M & 77.1 & 35.7 & 28.5 & 57.1 & 118.2 & 21.7 \\
        TR++ & 79.7M & 76.5 & 34.4 & 28.2 & 56.4  & 116.3 & 21.6 \\
        \bottomrule 
    \end{tabular}
    \normalsize
    \label{table:supp2}
\end{table}
\par
Regarding higher dropout rates for ZoDIAC,
Table \ref{table:supp1} shows the experiment results for ZoDIAC with sigmoid as scalar gate with higher dropout rates.
We observe that the performance of ZoDIAC with higher dropout rates is lower than the performance of ZoDIAC with the ones we found to be the best combination.
Also note that the version without GELU ($\dag$) achieves the worst performance.
\par
\begin{table}[!ht]\centering
    \footnotesize
    \caption {Results for ablations on grid search for higher dropout rates. Scalar gate is denoted as S-Gate. System and ZoDIAC dropout rates are denoted
    as S-Dr and Z-Dr and ZF denotes the Zoneup Factor and $\dag$ indicates that Gelu was removed throughout the whole system.}
    \begin{tabular}{ccccccccc}
        \toprule 
        \textbf
        & & & & & & {Metrics} \\
        S-Dr & Z-Dr & ZF & S-Gate & B1 & B4 & S & C \\
        \textbf{0.1} & \textbf{0.2} & 1.0 & $\sigma$ & \textbf{77.01} & \textbf{36.12} & \textbf{21.35} & \textbf{115.53} \\
        0.2$\dag$ & 0.3$\dag$ & 1.0 & $\sigma$ & 73.69 & 33.12 & 19.45 & 106.04 \\
        0.2 & 0.3 & 1.0 & $\sigma$ & 75.83 & 27.94 & 20.82 & 113.51 \\
        0.2 & 0.4 & 1.0 & $\sigma$ & 76.85 & 35.86 & 20.75 & 113.03 \\
        0.2 & 0.5 & 1.0 & $\sigma$ & 76.79 & 35.67 & 20.64 & 112.01 \\
        0.3 & 0.4 & 1.0 & $\sigma$ & 76.13 & 35.24 & 20.42 & 111.81 \\
        0.3 & 0.5 & 1.0 & $\sigma$ & 75.84 & 35.03  & 20.19 & 111.65 \\
        0.4 & 0.4 & 1.0 & $\sigma$ & 75.54 & 34.93  & 20.07 & 111.47 \\
        0.4 & 0.5 & 1.0 & $\sigma$ & 75.14 & 34.79  & 20.01 & 111.13 \\
        \bottomrule 
    \end{tabular}
    \normalsize
    \label{table:supp1}
\end{table}

\par
By looking at the ablation studies we can conclude that the performance gain of ZoDIAC is not due to the increase in parameters.
Also, we can conclude that for our task and domain, we found the dropout rates of 0.1 and 0.2 for system and ZoDIAC dropout rates to be the best combination,
however one can tune these hyper-parameters for their own task and domain.

%% file: sec6.tex
\section{Experiments with Convolutional Feature Extractor}
\par
\begin{table*}[ht!]\centering
    \caption {Results for experiments on MS-COCO Karpathy's test trained with XE loss.}
    \begin{tabular}{lccccc}
        \toprule 
        & & & \textbf{Metrics} \\
        \cline{2-6}
        Model & B1 & B4 & M & S & C \\ \hline
        ZoDIAC(tanh) & \textbf{77.4} & \textbf{36.6} & \textbf{28.0} & \textbf{21.4} & \textbf{115.9}\\
        ZoDIAC(sigmoid) & \textbf{77.0} & \textbf{36.1} & \textbf{27.9} & \textbf{21.3} & \textbf{115.5}\\
        Transformer(ours) & 75.9 & 35.0 & 27.6 & 21.0 & 114.0 \\ 
        Transformer\cite{Sharma_etal_2018} & 76.1 & 34.0 & 27.6 & 21.0 & 113.3 \\ 
        \hline
        ZoDIAC+AoA(tanh) & \textbf{77.4} & \textbf{36.8} & \textbf{28.2} & \textbf{21.6} & \textbf{117.6} \\ 
        ZoDIAC+AoA(sigmoid) & \textbf{77.3} & \textbf{36.8} & \textbf{28.2} & \textbf{21.6} & \textbf{117.4} \\ 
        AoA(ours) & 77.1 & 36.7 & 28.1 & 21.5 & 117.3 \\ 
        AoA\cite{Huang_2019_ICCV} & 77.4 & 37.2 & 28.4 & 21.3 & 119.8 \\ 
        \hline
        OFA\cite{wang2022OFA} & - & 43.9 & 31.8 & 24.8 & 145.3 \\
        SimVLM\cite{wang2022simvlm} & - & 40.6 & 33.7 & 25.4 & 143.3 \\
        LEMON\cite{LEMON} & - & 41.5 & 30.8 & 24.1 & 139.1 \\
     \bottomrule 
    \end{tabular}
    \label{table5}
\end{table*}
\begin{table*}[ht!]\centering
    \caption {Results for ensemble of 5 runs on MS-COCO Karpathy's test trained with XE loss.}
    \begin{tabular}{lccccc}
        \toprule 
        & & & \textbf{Metrics} \\
        \cline{2-6}
        Model & B1 & B4 & M & S & C \\ \hline
        ZoDIAC(tanh) & \textbf{78.5} & \textbf{38.4} & \textbf{28.8} & \textbf{22.0} & \textbf{121.4}\\
        ZoDIAC(sigmoid) & \textbf{78.1} & \textbf{38.3} & \textbf{28.6} & \textbf{21.9} & \textbf{121.3}\\
        Transformer(ours) & 77.2 & 37.2 & 27.6 & 21.7 & 119.1 \\
     \bottomrule 
    \end{tabular}
    \label{table6}
\end{table*}
\par
\begin{table*}[ht!]\centering
    \caption {Results for experiments on MS-COCO Karpathy's test trained with SCST loss.}
    \begin{tabular}{lccccc}
        \toprule 
        & & & \textbf{Metrics} \\
        \cline{2-6}
        Model & B1 & B4 & M & S & C \\ \hline
        ZoDIAC(tanh) & \textbf{80.6} & \textbf{38.7} & \textbf{29.1} & \textbf{22.7} & \textbf{129.6}\\
        ZoDIAC(sigmoid) & \textbf{80.4} & \textbf{38.7} & \textbf{29.0} & \textbf{22.7} & \textbf{129.5}\\
        Transformer(ours) & 80.2 & 38.6 & 28.9 & 22.6 & 129.2 \\ 
        Transformer\cite{Sharma_etal_2018} & 80.2 & 38.6 & 28.8 & 22.6 & 128.3 \\ 
        \hline
        OFA\cite{wang2022OFA} & - & 44.9 & 32.5 & 26.6 & 154.9 \\
        LEMON\cite{LEMON} & - & 42.6 & 31.4 & 25.5 & 145.5 \\
     \bottomrule 
    \end{tabular}
    \label{table7}
\end{table*}
\par
In the previous section we discussed the results of our experiments mainly using the CLIP models for visual feature extraction from regions of interest (RoIs).
Here in this section we offer more experiment results using convolutional feature extractors, mainly ResNet101 \cite{Resnet} trained on ImageNet \cite{imagenet_cvpr09}. This would show that the perfrmance gains of ZoDIAC model in comparison with the Transformer model is not a consequence of using CLIP based feature extractors that use attention mechanism and contrastive loss, as the ResNet101 model is purely convolutional and only is trained with cross-entropy loss rather than the contrastive loss as used in CLIP models.

\par
The results of our experiments on the MS-COCO trained with cross-entropy loss and ResNet101 as visual feature extractor are reported in Table \ref{table5}. These results are reported from best of 5 runs for each model.
Table \ref{table6} shows the results of ensemble evaluation using 5 runs.
We show the results for experiments using self-critical sequence training (SCST) in Table \ref{table7}.
We consider the CIDER score as the most important metric for all our experiments, although other metrics are indicators of performance as well, however CIDER
is the only metric that focuses on the quality of generated captions given an image and a set of ground truth captions.
This is also the main reason why we optimize the model on CIDER metric at the self-critical training stage.

Looking at the results from Table \ref{table5}, we observe that the results of ZoDIAC are superior to the self-attention module used inside the conventional Transformer model.
For the sake of comparison, we also report the results of the Transformer model on MS-COCO reported by Sharma et al. \cite{Sharma_etal_2018}.
It is worthy of being mentioned that when tanh is used as a scalar activation gate instead of the sigmoid ($\sigma$), the results are further improved.
We believe this is since the tanh activation gate creates a value between negative one and one, whereas the sigmoid function creates a value between 0 and 1,
By adding the resulting values to zoneout factor (set as 1), ZoDIAC is creating different refinement effects.
When tanh is used, the past intensity value can be less than 1 or greater than 1.
When sigmoid is used, considering the zoneout factor, the past intensity value is always between 1 and 2.
In other words, when tanh is used, the model can decrease or increase the intensity of attention values, whereas when sigmoid is used the model only has to learn
how much it should increase the intensity of attention values.
\par
Furthermore, we observe that when ZoDIAC is used as an extension inside another model that leverages self-attention,
namely the Attention-on-Attention (AOA) model \cite{Huang_2019_ICCV}, the results are also improved,
but the gains are marginal compared to when ZoDIAC is used inside the Transformer model.
Our investigations reveal that this is because AoA model employs an LSTM alongside multi-head attention and their proposed
refinement modules. In other words, there are fewer self-attention modules used inside AoA model, in comparison with the transformer model.
Likewise, our experiments reveal that ZoDIAC could be a potentially good choice as an extension for models that highly leverage on a self-attention module with minimal changes inside the Transformer model
and more changes in parameter size and number of layers.
Also in Table \ref{table5} we observe that AoA achieves slightly lower results than the ones reported by Sharma et al. \cite{Sharma_etal_2018}.
For the reasons discussed above, we did not perform further experiments with ZoDIAC the inside AoA model.
We also found that when ZoDIAC is extended to AoA model, we have to use a ZoDIAC dropout rate of $0.3$ (in order to match the properties of AoA) and also that we have to remove the GELU pre-activations before linear transformations are performed inside the multi-head attention module
in order to achieve better results than AoA.
\par
\par
By looking at the results from Table \ref{table6}, we can conclude that when trained with cross-entropy loss, ZoDIAC is a superior model for image captioning in
comparison with the conventional self-attention module inside the Transformer model. These results are reported from the ensemble evaluation of five models trained with different random
parameters. It is also worthy to mention that we could not find results reported in published literature from ensemble evaluation of the Transformer model on Karpathy's test set (MS-COCO dataset).
\par
For SCST loss training the results are reported in Table \ref{table7}. We can conclude that when trained with SCST loss, ZoDIAC achieves slightly better results in comparison
with the conventional self-attention module inside the Transformer model. Because performance gains for ZoDIAC+AoA model were marginal at the cross-entropy loss stage,
we decided not to run further experiments with ZoDIAC+AoA at the self-critical training stage.
We conclude that ZoDIAC can potentially be a good choice for achieving better results for models that leverage on self-attention when trained with SCST loss, however
these improvements are marginal compared to the improvements we achieve when the ZoDIAC model is trained with cross-entropy loss. Therefore, we encourage
the readers to investigate the effectiveness of ZoDIAC for tasks that do not require SCST loss training, such as image classification with transformers \cite{ViT}.
\par
At the current time, the state-of-the-art results in image captioning are generated by models such as OFA \cite{wang2022OFA}, LEMON \cite{LEMON} and SimVLM \cite{wang2022simvlm}
that utilize huge parameters and vision-language pre-training methods. For the OFA model, although the official code has been released by authors, the full pre-training data has not been released yet.
Also for LEMON and SimVLM, the official codes have not been released yet, therefore we could not conclude that ZoDIAC is a good choice for these models, but we encourage the readers
to investigate the effectiveness of ZoDIAC on these models or similar models that employ the self-attention module.
\par

%% file: sec7.tex
\section{Conclusions}\label{sec6}
\par
In this work, we proposed Zoneout Dropout Injection Attention Calculation (ZoDIAC), a novel attention mechanism that is a successor of
the self-attention mechanism in the Transformer model.
ZoDIAC leverages intensity value injection of attention map generated from values and second linear
projection of queries.
ZoDIAC improves the self-attention mechanism via refining the intra-relationships in the input sequence and injecting the intensities of the attention map generated
from the input sequence and another projection of it that can be considered as the secondary form of the input sequence.
We believe that ZoDIAC has opened the door to a new way of thinking about the self-attention mechanism in the Transformer model and how it could potentially
improve other models that utilize self-attention and we hope that
other members of the research community will benefit from this work and follow this research path.
\par